\documentclass[letterpaper, 10 pt, conference]{ieeeconf}  

\IEEEoverridecommandlockouts                              

\usepackage{cite}
\usepackage{amsmath,amssymb,amsfonts}
\usepackage{algorithmic}
\usepackage{graphicx}
\usepackage{textcomp}
\usepackage{xcolor}
\def\BibTeX{{\rm B\kern-.05em{\sc i\kern-.025em b}\kern-.08em
    T\kern-.1667em\lower.7ex\hbox{E}\kern-.125emX}}
    
\usepackage{amssymb}
\usepackage{pifont}
\usepackage[font=footnotesize,labelfont=bf]{caption}
\usepackage{multirow}
\usepackage{bm}
\usepackage{physics}
\usepackage{wrapfig}
\usepackage{makecell}
\usepackage[x11names, svgnames, dvipsnames]{xcolor}
\usepackage{color}

\usepackage{booktabs}
\usepackage{colortbl}

\newtheorem{ass}{\textbf{Assumption}}

\usepackage{tikz}
\usetikzlibrary{arrows.meta, positioning,shapes.geometric,calc, fit, backgrounds, decorations.pathreplacing,matrix,intersections}
\usepackage{graphicx}
\usepackage{pgfplots}
\pgfplotsset{compat=1.18}
\usepgfplotslibrary{fillbetween}
\definecolor{mediumelectricblue}{rgb}{0.01, 0.31, 0.59}
\definecolor{rufous}{rgb}{0.66, 0.11, 0.03}
\definecolor{venetianred}{rgb}{0.78, 0.03, 0.08}
\definecolor{royalblue}{rgb}{0.0, 0.14, 0.4}

\renewcommand{\eqref}[1]{Eq.~(\ref{#1})}
\usepackage[colorlinks=true, linkcolor=royalblue, urlcolor=royalblue, citecolor=royalblue]{hyperref}
\begin{document}

\title{\LARGE \bf
Multimodal Interaction Estimation for Contact-Aware \\Adaptive Impedance Control
}
\author{Mateusz Jaszczuk, Paris Perdikaris and Nadia Figueroa
\thanks{All authors are with the Department of Mechanical Engineering, University of Pennsylvania. Corresponding author: {\tt\small jaszczuk@engineering.upenn.edu}}
}
\maketitle

\begin{abstract}
With advancements in learning-based policies, robots are increasingly being deployed in human-centric environments, such as manufacturing floors and homes, where safe and precise operation requires compensating for unknown loads while remaining compliant during human interactions. Reliably distinguishing between these interaction types is challenging with a single sensing modality, as force signals cannot separate payloads from human pushes, and vision is susceptible to occlusions. We propose a \emph{task-agnostic} interaction estimation and adaptive control framework that addresses this challenge by fusing force history and 3D point cloud through a transformer-based classifier. Trained with modality dropout for occlusion resilience and fine-tuned from simulation to hardware, the classifier achieves 91.89\% accuracy and maintains above 90\% under complete occlusion of one modality. The classifier drives a system identification framework that selectively compensates for unknown payloads and a variable-impedance controller that switches between adaptive and reactive modes, while simultaneously estimating interaction forces for downstream tasks, such as human intent estimation. We validate the framework under external perturbations and unknown loads across single-arm manipulation, deployment with an imitation-learning-based policy, intent-driven target switching, and zero-shot bimanual deployment on two Franka arms lifting a box of unknown mass, demonstrating generalization to closed kinematic chains and unseen setups, enabling safe and precise deployment. Project website: \url{https://contact-aware-adaptive-impedance.github.io}
\end{abstract}

\vspace{0pt}
\section{Introduction}
Collaborative robots operating in human-centric environments must simultaneously maintain precise trajectory tracking and respond safely to human contact. While compliant behavior can be achieved through impedance control~\cite{Hogan1984} by reducing control gains or passivity-based techniques~\cite{Hogan1984,Stramigioli1999,Li1999,Duindam2004,Ferraguti2013,kronander2015passive,AlbuSchaffer2007,Kishi2003,Ferraguti2015,Energy2017,7801022,AbuDakka2020} these approaches rely on accurate knowledge of the robot's dynamics and struggle with unknown payloads and unmodeled actuator dynamics. Both of which degrade tracking performance and can result in unsafe behaviors. PID-based controllers can minimize tracking errors, but they sacrifice compliance to human touch. Humans, by contrast, seamlessly distinguish externally applied loads from contact perturbations, adjusting gravity compensation in real time~\cite{Riemann2002-aq,oh_speed_2021}. For example, when two coworkers collaboratively carry a heavy assembly part while remaining responsive to each other's pushes and redirection. Achieving this, however, requires resilience to visual occlusions and force ambiguity. To form more precise environmental and interaction estimates than either sensory channel alone, humans combine information across sensor modalities, for example, integrating visual and proprioceptive cues to estimate the position of one's hand while manipulating an object~\cite{proproception_vision}, or visual and haptic signals to judge object properties~\cite{haptics_vision}. Just as humans rely on both touch and vision to understand interactions, achieving interaction-awareness in robots requires multimodal perception to classify the sources of external forces, even under occlusions, thereby enabling selective load compensation without sacrificing responsiveness to human contact, which is essential for safe human-robot collaboration.
\begin{figure}[t!]
\centering
\input{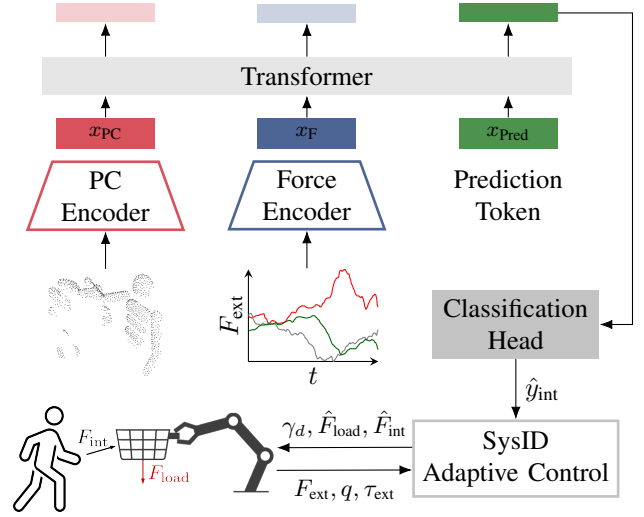}\caption{\label{fig:int_nn_framework}\textbf{Interaction Classification and Adaptive Control Framework.} We utilize a transformer-based architecture to fuse latent force and point cloud representations, allowing the model to reliably distinguish applied payloads from perturbations (applied to any link), allowing the adaptive controller to adjust gravity compensation and compliance accordingly.}
\vspace{-20pt}
\end{figure}

\textbf{Related Work} Physical human-robot interactions and precise impedance control require robots to reliably interpret contact events and adapt their behavior accordingly, which spans force estimation, interaction detection, and compliant control. Human contact detection from proprioception alone has been extensively studied, and typically requires several estimation-control layers and an accurate dynamics model~\cite{8059840}, limiting its robustness when the dynamics are unknown, for example, when manipulating an unknown payload. Adaptive human-robot interaction strategies have proposed using passive dynamical systems (DS)-based control ~\cite{Khoramshahi2019, Khoramshahi2020}, enabling intuitive human-guided control. More recently, \cite{doi:10.1126/scirobotics.adn4008} demonstrated that proprioception alone can enable an intrinsic robotic sense of touch, accurately localizing contact points on the robot's surface without tactile skin, enabling intuitive contact-based interaction. While effective, these methods respond to contact without explicitly classifying its source \emph{online}, for example, distinguishing whether an external force is due to a human push or to an attached load. External payload estimation methods~\cite{jaszczuk2025rapid, 6631072} enable robots to manipulate heavy external loads, but typically cannot reliably distinguish different types of forces, which is the core objective of this work. Further, contact-rich manipulation methods~\cite{9197509, LiuJ-RSS-25, 10801731, 8794219, 7759578} demonstrate effective force-aware policies but focus on objectives for learning motion policies rather than on understanding the source of interaction and estimating force.

\textbf{Contribution} We propose a \emph{task-agnostic} \textbf{adaptive control framework} for \emph{online} interaction classification and force estimation that distinguishes unknown payloads from human perturbations applied anywhere on the robot in real time and adapts gravity compensation and compliance accordingly. At its core, we developed a lightweight transformer-based multimodal classifier that fuses force history derived from proprioceptive sensing with a 3D point cloud, enabling an information-rich representation of the interaction (i.e., human perturbation or applied external load) experienced by the robot, thereby outperforming single-modality baselines by combining evidence from both inputs. To ensure robustness to sensor occlusion, we implemented modality dropout, enabling accurate classification from either modality alone, achieving over 90\% accuracy even with full occlusion of one sensor. To enhance generalization, we pretrain the network in simulation, then fine-tune it on a small real-robot dataset, achieving 91.89\% classification accuracy on real hardware, outperforming thresholding-based and MLP baselines, and generalizing to an unseen bimanual setup, and enabling zero-shot payload adaptation of an imitation learning policy not originally trained to handle loads. The classifier drives a system identification framework that selectively compensates for unknown loads and a variable-impedance controller that switches between adaptive and reactive modes, while estimating interaction forces for downstream tasks, such as human intent estimation.

\section{Background and Problem Formulation}\label{sec:problem_formulation}
Let us define the rigid-body dynamics of an $n$-DoF manipulator arm~\cite{spong2020},
\begin{equation} \label{eq:dynamics}
    M(q)\ddot{q}+C(q,\dot{q})\dot{q}+G(q) = \tau_c + \tau_{\text{ext}}
\end{equation}
\noindent where $\ddot{q},\dot{q},q \in\mathbb{R}^n$, represent joint accelerations, velocities, and positions, respectively, $M(q)\in\mathbb{R}^{n \times n}$, $C(q,\dot{q})\in\mathbb{R}^{n \times n}$, and  $G(q)\in\mathbb{R}^{n}$ represent the mass/inertia matrix, Coriolis matrix, and Gravity vector, respectively.  $\tau_c \in \mathbb{R}^n$ and $\tau_{\text{ext}} = J(q)^\top F_{\text{ext}}\in \mathbb{R}^n$ represent control torque, compute with impedance controller~\cite{Hogan1984}, and external torque at each joint, respectively, with $J(q)\in\mathbb{R}^{6\times n}$ being the Jacobian. 

To ensure passive interaction, we choose to track the task-space velocities from a DS-based motion planner $f(x)$ \cite{kronander2015passive}, learned offline or adapted online~\cite{dsbook}, by introducing the task-space variable impedance controller~\cite{shao2024constraintaware}
\begin{equation} \label{eq:passive_impedance}
    W_{c}(x)=G_x(q)-\gamma_d(\cdot) D(x)\big(\dot{x}-f(x)\big),~~~\tau_c=J(q)^\top W_c
\end{equation}
with task-space gravity vector $G_x(q)\in\mathbb{R}^6$, damping matrix $D(x)\succ 0\in\mathbb{R}^{6\times 6}$, desired velocity from $f(x)$, and variable impedance parameter $\gamma_d(\cdot)>0$, which is conditioned on the current classified interaction from our estimation framework.
\begin{ass}
    While the classifier can detect contacts at any link, we estimate all forces w.r.t. the EE, as manipulators primarily interact with the environment through it. Thus, we assume that external torque $\tau_\text{ext}$ in \eqref{eq:dynamics} is caused by either an unknown payload or an external force (perturbation or collision) applied anywhere to the robot's structure, both expressed in the EE frame as
    \begin{equation}
        \tau_\text{ext} = J(q)^{\top}\begin{bmatrix}
         \textcolor{venetianred}{F_\text{load}} + \textcolor{royalblue}{F_\text{int}}\\ r_{\text{CoM} \times (F_\text{load}+F_\text{int})}
     \end{bmatrix}
    \end{equation}
    where $\textcolor{royalblue}{F_\text{int}}$ is an external interaction force applied to the EE and $\textcolor{venetianred}{F_\text{load}}$ is unknown payload force.
\end{ass}

\textbf{Problem} The performance of impedance controllers relies on accurate system dynamics equations, as any unknown payload will result in uncompensated $\tau_\text{ext}$, compromising tracking performance \cite{8059840}. While existing frameworks~\cite{jaszczuk2025rapid, provably_safe_sys_id, Srour2025} can adapt to unknown loads attached to the end-effector, they are not designed to disambiguate load and interaction-related forces, leading to unsafe behaviors when the two coincide. Further, while impedance controllers allow compliant responses to perturbations, they require low impedance gains to achieve this, which degrades tracking. Therefore, to allow precise execution and compliance, we need to distinguish $\textcolor{venetianred}{F_\text{load}}$ from $\textcolor{royalblue}{F_\text{int}}$ and compensate only for $\textcolor{venetianred}{F_\text{load}}$ to avoid instabilities and poor tracking. 

\textbf{Goal} Therefore, we aim to classify interaction events as either \textcolor{venetianred}{\emph{unknown payloads}} or \textcolor{royalblue}{\emph{perturbations}} in real time, compensating for unknown payloads and measuring effective interaction force $(\textcolor{venetianred}{F_\text{load}},\textcolor{royalblue}{F_\text{int}})$. We further want to use the interaction estimation to actively adjust the controller's impedance  $\gamma_d(\cdot)$, enabling precise tracking in reactive mode and compliant behavior when human perturbation is detected. 
\section{Interaction Classifier}
\label{sec:int_classifier}
To distinguish human perturbations $(\textcolor{royalblue}{F_\text{int}})$ from external loads $(\textcolor{venetianred}{F_\text{load}})$, we design a transformer-based classifier that fuses proprioceptive and visual inputs to estimate the current interaction, inspired by transformer-based multimodal fusion approaches in robotics~\cite{haldar2024baku, chen2022visuotactile, Brohan-RSS-23, 10611293}. 
By aggregating information from both modalities, the framework maintains high accuracy even when an interaction is not fully observable from a single modality, i.e., due to occlusion or ambiguous forces.
\vspace{-2.5pt}
\subsection{Architecture}
\label{sec:int_classifier_architecture}
Our classifier (Fig.~\ref{fig:int_nn_framework}) consists of two modality encoders, producing compact latent tokens in $\mathbb{R}^{64}$, which, along with a prediction token, are fused by our novel transformer (Fig.~\ref{fig:transformer}) into a latent multimodal representation for classification\footnote{\label{website}Further architectural, training details and analyses can be found in supplementary project website \url{https://contact-aware-adaptive-impedance.github.io}}.
\subsubsection{Force History Encoder}\label{sec:F_enc} To process the $F_{\text{ext}}$ history, we designed a force encoder that produces a compact latent force representation for the transformer module. The force history $F_\text{ext}\in \mathbb{R}^{25 \times 3}$ ($\sim250$\,ms) is passed through a two-layer 1D Convolution Network~\cite{LeCun_cnn} with channel dimensions $3\rightarrow 32 \rightarrow32$ and kernel size $5$, followed by attention pooling~\cite{attention} for adaptive feature extraction, and an MLP with architecture $32 \rightarrow 16 \rightarrow 64$ and ReLU activation. Finally, we apply an output projection and Layer Normalization~\cite{ba2016layernormalization} to extract the latent force representation, $x_{F}\in \mathbb{R}^{64}$.

\subsubsection{Point Cloud Encoder} To process visual information, we employ a Point Cloud encoder~\cite{Ze2024DP3} that extracts a latent 3D representation of the scene. A raw point cloud is cropped and furthest-point sampled to $\mathbb{R}^{1024 \times 3}$, then passed through the MLP with an architecture of $3 \rightarrow 64 \rightarrow 128 \rightarrow 256$ and ReLU activations, followed by max pooling to aggregate global geometric features. Finally, we apply output projection and Layer Normalization~\cite{ba2016layernormalization} to ensure the latent 3D representation, $x_{\text{PC}}\in \mathbb{R}^{64}$, is compatible in scale and dimension with the representations produced by other encoders. 

\subsubsection{Transformer}
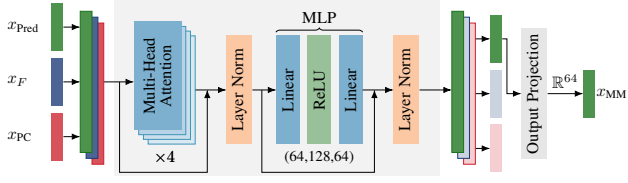
\begin{figure}
\centering
\resizebox{\columnwidth}{!}{

\begin{tikzpicture}[
      block/.style  = {rectangle, minimum width=2.5cm, minimum height=0.75cm, align=center},
      text_box/.style      = {rectangle, minimum width=3cm, minimum height=0.25cm, align=left},
      text_boxin/.style      = {rectangle, minimum width=0.75cm, minimum height=0.5cm,text width=1cm},
      Attlayer/.style      = {rectangle, minimum width=1cm, minimum height=2.3cm},
      layer/.style      = {rectangle, minimum width=0.5cm, minimum height=2.3cm},
      token/.style = {rectangle, minimum width=0.25cm, minimum height=1cm},
      tokentall/.style = {rectangle, minimum width=0.25cm, minimum height=3cm},
      >=Latex
    ]

    \foreach \i/\pct in {3/10,2/15,1/20} {
      \node[Attlayer, draw=MidnightBlue!60, line width=0.4pt, fill=MidnightBlue!\pct]
        (SelfAttBack\i) at (\i*0.1,-\i*0.1) {};}
    \node[Attlayer, fill=MidnightBlue!40]
      (SelfAttTop) at (0,0)
      {\rotatebox{90}{\shortstack{Multi-Head\\Attention}}};
    \node[below=5pt]at ($(SelfAttTop.south)!0.5!(SelfAttBack3.south)$){$\times 4$};

    \node[layer, fill=Orange!40, right=0.65cm of SelfAttBack3](layerLn1){\rotatebox{90}{Layer Norm}};

    \node[layer, fill=MidnightBlue!40, right=0.5cm of layerLn1.east](layerLin2){\rotatebox{90}{Linear}};
    \node[layer, fill=DarkGreen!40, right=0.15cm of layerLin2]  (layerReLu2){\rotatebox{90}{ReLU}};
    \node[layer, fill=MidnightBlue!40, right=0.15cm of layerReLu2](layerLin3){\rotatebox{90}{Linear}};
    \node[text_box, below=0.01cm of  layerReLu2](MLP_text_layers) {\small{(64,128,64)}};
    \draw[decorate, decoration={brace, raise=2pt}](layerLin2.north west) --(layerLin3.north east) node[midway, above=4pt](underbrace_mlp) {MLP};
    \node[layer, fill=Orange!40, right=0.65cm of layerLin3](layerLn4){\rotatebox{90}{Layer Norm}};
    \begin{scope}[on background layer]
        \node[rectangle, fill=gray!10, inner xsep=12pt,   
        inner ysep=4pt,   fit=(layerLin2)(underbrace_mlp)(layerReLu2)(MLP_text_layers)(SelfAttTop)(layerLn4)
    ](Transfromer){};
    \end{scope}

    \node[tokentall, draw=venetianred!90,fill=venetianred!70, left=0.2cm of Transfromer, yshift=-0.125cm](tokenPCs){};
    \node[tokentall, draw=royalblue!90, line width=0.4pt, fill=royalblue!70]
    at ($(tokenPCs.center)-(0.125cm,-0.125cm)$) (tokenFs) {};
    \node[tokentall,draw=DarkGreen!90, fill=DarkGreen!70]
    at ($(tokenFs.center)-(0.125cm,-0.125cm)$) (tokenPreds) {};
    \begin{scope}[on background layer]
        \draw[->] (tokenPreds.east)    -- (tokenPreds.east -| SelfAttTop.west);

    \end{scope}

    \node[token, fill=royalblue!70, left=0.35cm of tokenPreds](tokenF){};
    \node[token, fill=venetianred!70, below=0.15cm of tokenF](tokenPC){};
    \node[token, fill=DarkGreen!70, above=0.15cm of tokenF](tokenPred){};

    \node[text_boxin, align=left, left=-0.2cm of tokenPC.west] (textPC) {$x_{\text{PC}\phantom{red}}$};

    \node[text_boxin,  align=left, left=-0.2cm of tokenF.west] (textF) {$x_{F\phantom{pred}}$};

    \node[text_boxin, align=left, left=-0.2cm of tokenPred.west] (textPred) {$x_{\text{Pred}}$};
    
    \draw[->] (tokenPC.east)    -- (tokenPC.east -| tokenPreds.west);
    \draw[->] (tokenF.east)    -- (tokenF.east -| tokenPreds.west);
    \draw[->] (tokenPred.east)    -- (tokenPred.east -| tokenPreds.west);

    \node[tokentall, draw=venetianred!90,fill=venetianred!20, right=0.5cm of Transfromer, yshift=-0.125cm](tokenPCsout){};
    \node[tokentall, draw=royalblue!90, line width=0.4pt, fill=royalblue!20]
    at ($(tokenPCsout.center)-(0.125cm,-0.125cm)$) (tokenFsout) {};
    \node[tokentall,draw=DarkGreen!90, fill=DarkGreen!70]
    at ($(tokenFsout.center)-(0.125cm,-0.125cm)$) (tokenPredsout) {};
    \begin{scope}[on background layer]
        \draw[->] (layerLn4.east)    -- (layerLn4.east -| tokenPredsout.west);

    \end{scope}

    \node[token, fill=royalblue!20, right=0.3cm of tokenPCsout](outtokenF){};
    \node[token, fill=venetianred!20, below=0.15cm of outtokenF](outtokenPC){};
    \node[token, fill=DarkGreen!70, above=0.15cm of outtokenF](outtokenPred){};

    \draw[->]  (outtokenPC.west -| tokenPCsout.east)   --(outtokenPC.west) ;
    \draw[->]  (outtokenPred.west -| tokenPCsout.east)   --(outtokenPred.west) ;
    \draw[->]  (outtokenF.west -| tokenPCsout.east)   --(outtokenF.west) ;

    \node[layer, fill=gray!20, right=0.4cm of outtokenF.east](OutputProj){\rotatebox{90}{Output Projection}};

    \node[token, fill=DarkGreen!70, right=0.75cm of OutputProj.east](outToken){};
    \node[text_boxin, align=left, right=-0.1cm of outToken.east] (textOut) {$x_{\text{MM}}$};
    \node[below=5pt]at ($(SelfAttTop.south)!0.5!(SelfAttBack3.south)$){$\times 4$};
    \draw[->](OutputProj.east) -- node[above]{$\mathbb{R}^{64}$}(outToken.west);
    \coordinate (midA) at ($(outtokenPred.east)+(0.1cm,0)$);

    \draw[->]
    (outtokenPred.east) -- (midA) -- (midA |-OutputProj.west)-- (OutputProj.west);

    \coordinate (targetA) at ($(layerLn1.west)+(-0.4cm,-1.75cm)$);
    \coordinate (p0) at ($(tokenPreds.east -| SelfAttTop.west)+(-0.3cm,0)$);
    
    \draw[->]
      (p0)
      -- (p0 |- targetA) 
      -- (targetA)
      -- ($(layerLn1.west)+(-0.4cm,0cm)$);
    \draw[->](SelfAttBack3.east) 
      -- (layerLn1.west);

      \coordinate (targetB) at ($(layerLn4.west)+(-0.4cm,-1.75cm)$);
    \coordinate (p1) at ($(layerLin3.east -| layerLin2.west)+(-0.3cm,0)$);
    
    \draw[->]
      (p1)
      -- (p1 |- targetB) 
      -- (targetB)
      -- ($(layerLn4.west)+(-0.4cm,0cm)$);
    \draw[->](layerLin3.east) 
      -- (layerLn4.west);
      
    \draw[->](layerLn1.east) -- (layerLin2.west);

\end{tikzpicture}}
\caption{\label{fig:transformer}Multimodal Transformer architecture.}
\vspace{-12pt}
\end{figure}
To learn latent multimodal interaction representations, we design a transformer-based architecture~\cite{attention} that integrates visual and proprioceptive cues (Fig~\ref{fig:transformer}). The transformer receives the Force, the Point Cloud, and learnable Interaction Classification Tokens, which are jointly passed through a Multi-Head Self-Attention (4 heads) to model cross-modal dependencies, followed by Layer Normalization~\cite{ba2016layernormalization}, an MLP with architecture $64 \rightarrow 128 \rightarrow 64$ with ReLU activations, and a second Layer Normalization, with Residual connections around both attention and MLP modules. The learnable prediction token, having attended to both modalities, is then extracted from the transformer output and passed through the output projection to yield the latent multimodal interaction representation\footnote{Note that, if unimodal latent representations are not needed, the attention module can be replaced with cross attention, where the classification token attends directly over the three stacked input tokens.}, $x_\text{MM}\in \mathbb{R}^{64}.$

\subsubsection{Classification Head} We pass the interaction representation $x_\text{MM}\in \mathbb{R}^{64}$ through an MLP with architecture $64 \rightarrow 128 \rightarrow 256 \rightarrow 1$ and ReLU activations, followed by a sigmoid function and max pooling to produce the final interaction classification prediction, $\hat{y}_\text{int}$.

\subsection{Training}\label{sec:int_nn_traning}
\subsubsection{Simulation-Based Pretraining} 
\begin{figure}
    \centering
    \resizebox{0.95\linewidth}{!}{\input{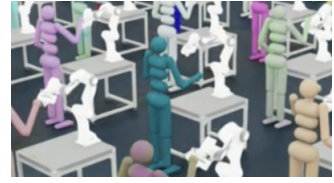}}
    \caption{IsaacLab dataset generation and domain randomization.\label{fig:domain_rand}}
    \vspace{-5pt}
\end{figure}

To generate a diversified dataset for model pretraining, we used IsaacLab~\cite{mittal2025isaaclab} with a Franka manipulator and humanoid model, as shown in Fig.~\ref{fig:domain_rand}. During a simulation, the humanoid perturbs the robot with random contact points, forces, and durations (Fig.~\ref{fig:domain_rand}), while the robot is in a random configuration and is subject to a random external load applied to the end-effector. This domain randomization enables the recording of a diverse dataset of 15,000 samples (approx. 208 minutes of operation), capturing various interaction scenarios. Encoders were pretrained with lr=$10^{-4}$ (5000/1000 epochs for Force/PC encoders), followed by full model pretraining with encoders lr=$10^{-5}$ and transformer/head  lr=$10^{-4}$ (1000 epochs).

\subsubsection{Fine-Tuning with Real Robot Demonstrations} To close the sim-to-real gap, we collected an additional 25,000 training samples from approximately 6.75 minutes of real-world robot deployment, including manipulation under unknown loads and perturbations, and fine-tuned the network on a dataset containing both simulation and real-robot data (10,000/25,000). Model was fine-tuned with lr=$5\times10^{-5}$ for 1000 epochs.

\subsubsection{Overfitting Mitigation and Occlusion Resilience} To prevent the model from over-reliance on a single modality, we implemented modality dropout. During training, a random subset of the force measurements or point cloud input is occluded and replaced with Gaussian noise. This encourages it to extract meaningful features from both sensory inputs and to improve its reliability under sensory occlusions. To further prevent overfitting to the training data, we implemented dropout~\cite{dropout} in the force encoder, the transformer block (self-attention and MLP layers), and the model head\footnote{For training, we utilized PyTorch Library~\cite{paszke2019pytorchimperativestylehighperformance} and workstation with an Intel\textsuperscript{\textregistered} Core\texttrademark{} Ultra~9~285K CPU, NVIDIA\textsuperscript{\textregistered} GeForce RTX\texttrademark{} 5090 GPU.}.  

\begin{figure}[!tbp]
\centering
\resizebox{1.0\linewidth}{!}{
\begin{tikzpicture}
\pgfmathsetmacro{\a}{1816}
\pgfmathsetmacro{\b}{29}
\pgfmathsetmacro{\c}{225}
\pgfmathsetmacro{\d}{930}
\pgfmathsetmacro{\TotalCount}{\a+\b+\c+\d}
\pgfmathtruncatemacro{\pa}{100*(\a/(\TotalCount))}
\pgfmathtruncatemacro{\pb}{100*(\b/(\TotalCount))}
\pgfmathtruncatemacro{\pc}{100*(\c/(\TotalCount))}
\pgfmathtruncatemacro{\pd}{100*(\d/(\TotalCount))}
\matrix (cm_force) [matrix of nodes,row sep=-\pgflinewidth,
  column sep=-\pgflinewidth,
  nodes={draw,minimum size=1.3cm, align=center, font=\small}]{
  |[fill=gray!\pa]| \a & |[fill=gray!\pb]| \b \\
  |[fill=gray!\pc]| \c &  |[fill=gray!\pd]| \d \\
};
\node[below=0.0cm of cm_force-2-1] {\scriptsize{Reactive}};
\node[below=0.0cm of cm_force-2-2] {\scriptsize{Compliant}};
\node[left=0.07cm of cm_force-1-1] {\rotatebox{90}{\scriptsize{Reactive}}};
\node[left=0.0cm of cm_force-2-1] {\rotatebox{90}{\scriptsize{Compliant}}};
\node[below=0.2cm of cm_force] {\small Predicted};
\node[left=0.25cm of cm_force] {\rotatebox{90}{\small True}};
\node[above=-0.1cm of cm_force](fm_acc){\small Accuracy: $91.5 \%$};
\node[above=-0.1cm of fm_acc] {\textbf{Force Model}};
 
\pgfmathsetmacro{\a}{1728}
\pgfmathsetmacro{\b}{63}
\pgfmathsetmacro{\c}{63}
\pgfmathsetmacro{\d}{1092}
\pgfmathsetmacro{\TotalCount}{\a+\b+\c+\d}
\pgfmathtruncatemacro{\pa}{100*(\a/(\TotalCount))}
\pgfmathtruncatemacro{\pb}{100*(\b/(\TotalCount))}
\pgfmathtruncatemacro{\pc}{100*(\c/(\TotalCount))}
\pgfmathtruncatemacro{\pd}{100*(\d/(\TotalCount))}
\matrix (cm_pc) [matrix of nodes, row sep=-\pgflinewidth,
  column sep=-\pgflinewidth, right=0.05cm of cm_force,
  nodes={draw,minimum size=1.3cm, align=center, font=\small}]{
  |[fill=gray!\pa]| \a & |[fill=gray!\pb]| \b \\
  |[fill=gray!\pc]| \c &  |[fill=gray!\pd]| \d \\
};
\node[below=0.0cm of cm_pc-2-1] {\scriptsize{Reactive}};
\node[below=0.0cm of cm_pc-2-2] {\scriptsize{Compliant}};
\node[below=0.2cm of cm_pc] {\small Predicted};
\node[above=-0.1cm of cm_pc](pc_acc){\small Accuracy: $95.8 \%$};
\node[above=-0.1cm of pc_acc] {\textbf{PC Model}};
 
\pgfmathsetmacro{\a}{1817}
\pgfmathsetmacro{\b}{28}
\pgfmathsetmacro{\c}{50}
\pgfmathsetmacro{\d}{1105}
\pgfmathsetmacro{\TotalCount}{\a+\b+\c+\d}
\pgfmathtruncatemacro{\pa}{100*(\a/(\TotalCount))}
\pgfmathtruncatemacro{\pb}{100*(\b/(\TotalCount))}
\pgfmathtruncatemacro{\pc}{100*(\c/(\TotalCount))}
\pgfmathtruncatemacro{\pd}{100*(\d/(\TotalCount))}
\matrix (cm_mm) [matrix of nodes, row sep=-\pgflinewidth,
  column sep=-\pgflinewidth, right=0.05cm of cm_pc,
  nodes={draw, minimum size=1.3cm, align=center, font=\small}]{
  |[fill=gray!\pa]| \a & |[fill=gray!\pb]| \b \\
  |[fill=gray!\pc]| \c &  |[fill=gray!\pd]| \d \\
};
\node[below=0.0cm of cm_mm-2-1] {\scriptsize{Reactive}};
\node[below=0.0cm of cm_mm-2-2] {\scriptsize{Compliant}};
\node[below=0.2cm of cm_mm] {\small Predicted};
\node[above=-0.1cm of cm_mm](f_acc){\small Accuracy: $97.4 \%$};
\node[above=-0.1cm of f_acc] {\textbf{Multimodal Model}};
\end{tikzpicture}}
\caption{\label{fig:sim_confusion_mat}Confusion matrices for simulation-trained classifiers evaluated on a simulation for the force-only, point cloud-only, and multimodal models.}
\vspace{-10pt}
\end{figure}

\subsection{Classifier Evaluation}\label{sec:int_nn_eval}
\subsubsection{Pretraining Classification Accuracy}
We assess the impact of joint multimodal representation by comparing to single-modality models with the same encoder and classification-head architecture. As shown in Fig.~\ref{fig:sim_confusion_mat}, the multimodal model outperforms single-modality baselines, producing fewer \emph{"false reactives"} where perturbations go undetected, enabling more robust performance.

\subsubsection{Modality Dropout Ablations} To assess robustness to sensor occlusion, we replace force and point cloud inputs with Gaussian noise and compare models trained with and without modality dropout. While both nominally achieved high accuracy (Fig.~\ref{fig:sim_confusion_mat}), dropout-free models fall below 50\% accuracy under occlusions. With modality dropout, accuracy recovers to above 90\% (Table~\ref{tab:ablation_md}), confirming that the model learns to classify interaction from either modality alone.

\begin{table}[!tbp]
\renewcommand{\arraystretch}{1.25}
\centering
\caption{Model classification accuracy under modality dropout ablations.}
\label{tab:ablation_md}
\begin{tabular}{ccc}
\toprule
 &\textbf{ w/o Modality Dropout }& \textbf{w/ Modality Dropout} \\ \hline
 Force Occlusion & 43.10\% & \cellcolor{green!10} \textbf{95.40\%} \\ 
 PC Occlusion & 38.53\% & \cellcolor{green!10} \textbf{90.10\%} \\ \bottomrule
\end{tabular}
\vspace{-5pt}
\end{table}

\subsubsection{Real-Robot Evaluation Dataset}\label{sec:robotdataset}
To assess the model's ability to continuously classify interaction events during deployment, we collected a diverse dataset from 3 participants interacting with the robot from various relative positions, using right/left hands arbitrarily to apply perturbations to the robot with varying length and strength at different parts of the robot (from link 1 to the EE). We repeated these trials under varying loads ($0-1.2$\,kg) with different fixed-gain controllers ($\gamma_d = 1.0-2.4$) tracking an equilibrium point, $f(x) = -(x-x^*)$, to capture the model's behavior across a wide range of conditions. Additionally, to verify that the classifier is not overfitting to the point cloud representation, we collected data where participants' hands were in close proximity to the robot; however, without perturbing it. As summarized in Table~\ref{tab:eval_dataset}, we collected over one hour of experimental data across six categories. Samples were collected every 80 ms and subsequently divided into N testing intervals of at most 80 s each, allowing us to compute experimental statistics across approximately uncorrelated intervals rather than raw data points (std. in Table~\ref{tab:updated_results} are computed based on N-interval-averaged accuracies).

\subsubsection{Sim-to-Real Transfer via Fine-Tuning} 
Without the fine-tuning stage, the simulation-trained model achieves \textbf{38.14\% accuracy (sim-only)} when evaluated on the Real-Robot Dataset, as unmodeled sensor noise causes it to over-predict perturbations (false positives). Fine-tuning on a mixed dataset of simulated and real data increases to \textbf{91.89\% accuracy (sim+real)}, enabling robust sim-to-real transfer while preserving the generalization benefits of pretraining domain randomization. For comparison, training on the real data alone yields \textbf{91.42\% accuracy (real-only)}, suggesting that sim pretraining is not strictly necessary to reach strong performance. However, since simulated data is inexpensive to generate, we believe that the sim+real approach still offers a favorable trade-off, and the modest accuracy gain over real-only training suggests it may provide some additional benefits during deployment.



\begin{table}[!t]
\vspace{-5pt}
\renewcommand{\arraystretch}{1.25}
\centering
\caption{Real-Robot Evaluation Dataset}
\label{tab:eval_dataset}
\vspace{-2.5pt}
\begin{tabular}{cccc}
\toprule
 \textbf{Experiment} & \textbf{Time (min)} & \textbf{Samples} & \textbf{Test Intervals}  \\ 
\hline
\rowcolor{gray!10} Load & 8.87 & 6,537  &7 \\
Perturbation & 28.35 & 20,864  & 21 \\
\rowcolor{gray!10} Perturbation (z) & 3.33& 2,457  & 3\\
Load + Pert. & 10.46& 7,618  & 8\\
\rowcolor{gray!10} Load + Pert (z) &7.47&5,519  & 6 \\
Contact Prox.& 5.90 & 4,301  & 5\\ 
\bottomrule
\textbf{Total}& \textbf{64.38} &\textbf{47,296} & \textbf{50}\\
\bottomrule
\end{tabular}
\vspace{-10pt}
\end{table}

\subsubsection{Real-Robot Evaluation}
We evaluated the classifier on the physical Franka arm\footnote{To evaluate the model with real-time control loops on the Franka manipulator, we used a workstation with an Intel\textsuperscript{\textregistered} Core\texttrademark{} i7-11700K CPU and an NVIDIA\textsuperscript{\textregistered} GeForce RTX\texttrademark{} 3070 GPU, and RoBoStack~\cite{FischerRAM2021} to establish software-controller-robot communications with ROS 1 Noetic.} across six experimental scenarios (Section~\ref{sec:robotdataset}) designed to test different aspects of interaction detection. Compared to the baseline (a modified version\footnote{The standard RME detection algorithm~\cite{jaszczuk2025rapid} was designed to detect loads rather than perturbations; we modified it to match the specifications of our proposed method for reliable comparison.} of the thresholding-based RME detection algorithm~\cite{jaszczuk2025rapid}), our approach improves in total classification accuracy from 80.66\% to 91.89\%, with the largest improvements in complex interactions, in particular when a perturbation is applied along the $z$-axis (\emph{Perturbation (z)} and \emph{Load + Pert (z)}) and produces a force that closely resembles the load, making the two difficult to distinguish for the baseline (using only force). Our model's joint representation of force and point cloud enables it to resolve this ambiguity by incorporating spatial contact information, rather than relying on force alone.
\begin{table}[!t]
\renewcommand{\arraystretch}{1.25}
\centering
\caption{Classification accuracy across various real-robot experimental scenarios for RME-based Baseline~\cite{jaszczuk2025rapid}, Multimodal-MLP and  -Transformer.}
\label{tab:updated_results}
\resizebox{\columnwidth}{!}{
\begin{tabular}{cccc}
\toprule
 \textbf{Experiment} & \textbf{Baseline}~\cite{jaszczuk2025rapid} &\textbf{MM-MLP} &\textbf{MM-Transformer} \\ \hline
 
\rowcolor{gray!10} Load & 100$\pm$0.00\% &99.74$\pm$0.76\% & 99.48$\pm$0.81\%\\

Perturbation & 82.18$\pm$10.62\%& 87.96$\pm$5.59\% & 87.87$\pm$5.26\%\\

\rowcolor{gray!10} Perturbation (z) & 40.33$\pm$6.92\%& 84.41$\pm$9.62\%&90.92$\pm$5.94\%\\

Load + Pert. &  81.58$\pm$6.53\%& 92.15$\pm$4.27\%& 91.65$\pm$3.44\%\\ 

\rowcolor{gray!10} Load + Pert (z)& 53.63$\pm$8.96\%& 94.08$\pm$2.89\%& 95.33$\pm$1.48\%\\

Contact Prox.& 100$\pm$0.00\%& 97.88$\pm$1.85\% & 96.44$\pm$3.59\%\\
\bottomrule
\textbf{Total} & \textbf{80.66\%}& \textbf{91.70\%} &\cellcolor{green!10}\textbf{91.89\%} \\
\bottomrule
\end{tabular}}
\vspace{-5pt}
\end{table}

\begin{figure}[!tbp]
\centering
\begin{tikzpicture}[
      layer/.style  = {rectangle, minimum width=0.5cm, minimum height=2.3cm}
      >=Latex
    ]
   
    \node[align=center] (plot1) {
    \begin{tikzpicture}
    \begin{axis}[width=8cm, height=3.5cm, axis lines=left, xtick distance=10,
      ytick distance=15, xlabel={$t\,(\mathrm{s})$},
ylabel={$F_{\text{ext}}\,(\mathrm{N})$}, enlargelimits=false, ymin=-37.5, ymax=25.5,
      legend style={at={(1.02,0.5)}, anchor=west,draw=none,font=\small}, legend cell align=left]

    \addplot[smooth, thin, red]
    table[header=false, col sep=comma, x index=0, y index=1
    ]{TikZ_plots/test_plot.csv};

    \addplot[smooth, thin, black]
    table[header=false, col sep=comma, x index=0, y index=2
    ]{TikZ_plots/test_plot.csv};

    \addplot[smooth, thin, DarkGreen]
    table[header=false, col sep=comma, x index=0, y index=3
    ]{TikZ_plots/test_plot.csv};
  
    \def\barY{-100}
    \def\barH{200}
    \addplot[name path=baseGT, draw=none]
    table[col sep=comma, x index=0, y expr=\barY]
    {TikZ_plots/test_plot.csv};
    \addplot[name path=topGT, draw=none, const plot]
    table[col sep=comma, x index=0, y expr=\barY + (\thisrowno{6} > 0.5 ? \barH : 0)]{TikZ_plots/test_plot.csv};
    \addplot[draw=none, fill=orange!30, fill opacity=0.4, forget plot]
    fill between[of=baseGT and topGT];
    
    \def\barY{-100}
    \def\barH{200}
    \addplot[name path=baseGT, draw=none]
    table[col sep=comma, x index=0, y expr=\barY]
    {TikZ_plots/test_plot.csv};
    \addplot[name path=topGT, draw=none, const plot]
    table[col sep=comma, x index=0, y expr=\barY + (\thisrowno{4} > 0.5 ? \barH : 0)]{TikZ_plots/test_plot.csv};
    \addplot[draw=none, fill=gray!50, fill opacity=0.7, forget plot]
    fill between[of=baseGT and topGT];

    \def\barY{16}
    \def\barH{4}
    \addplot[name path=baseGT, draw=none]
    table[col sep=comma, x index=0, y expr=\barY]
    {TikZ_plots/test_plot.csv};
    \addplot[name path=topGT, draw=none, const plot]
    table[col sep=comma, x index=0, y expr=\barY + (\thisrowno{5} > 0.5 ? \barH : 0)]{TikZ_plots/test_plot.csv};
    \addplot[draw=none, fill=venetianred!70, fill opacity=0.9, forget plot]
    fill between[of=baseGT and topGT];
    \end{axis}
    \end{tikzpicture}%
    };
\end{tikzpicture}
\vspace{-10pt}
\caption{\label{fig:nn_robot_eval}Classifier's performance in continuous perturbation detection. Our model reliably flags contact (\textcolor{venetianred!90}{red}) during the intervals when perturbations (\textcolor{gray}{gray}) and loads (\textcolor{orange!60}{orange}) are applied, including cases where a $z$-axis perturbation is applied alongside loads. The \textcolor{red}{red}, black, and \textcolor{DarkGreen}{green} plots represent $F_{\text{ext}}$ in the x-, y-, and z-directions, respectively.}
\vspace{-0.6cm}
\end{figure}
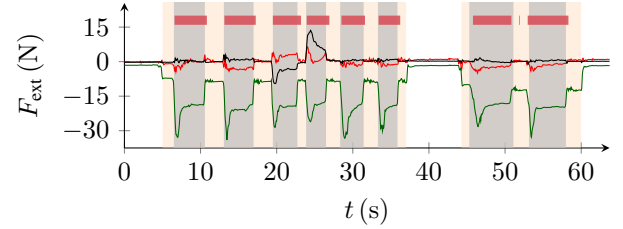

To assess the contribution of transformer-based feature aggregation, we also compare against MM-MLP, a variant of our framework that replaces the transformer with mean pooling over the same multimodal features. While the overall improvement of MM-Transformer is modest, there is consistent gain in the more \textit{ambiguous interaction} scenarios (\emph{Perturbation (z)} and \emph{Load + Pert (z)}), suggesting that the transformer's ability to adaptively weight features becomes valuable when interactions become more complex. Given that the average inference time of the transformer-based model is only $1.6$\,ms, it incurs negligible additional computational cost, enabling real-time deployment and supporting the use of the lightweight transformer over simple mean-pooling. 

The only cases in which the baseline slightly outperforms our model are \emph{Load} and \emph{Contact Proximity}, for both of which our accuracy still remains high, indicating no meaningful degradation on the simpler cases. Fig.~\ref{fig:nn_robot_eval} illustrates classification output while tracking an equilibrium point, $f(x) = -(x-x^*)$ subject to perturbations and loads.
\section{Adaptive Impedance Controller}\label{sec:ada_control}
The adaptive impedance controller uses the interaction-classification output, $\hat{y}_\text{int}\in[0,1]$, to (i) set the controller's compliance and to (ii) compensate for unknown loads via the RME's model-mismatch estimate~\cite{jaszczuk2025rapid}. Because the classification is stochastic, we use the most conservative (compliant) of the last 250 predictions\footnote{\label{fn:k250}{At the NN's inference rate of 1.6\,ms, 250-sample window spans $\sim400$\,ms interval, which exceeds both the RME-related delay~\cite{jaszczuk2025rapid} and classifier's detection delay ($\sim250$\,ms due to due to feedback window analysis, Section~\ref{sec:F_enc}), thus allowing safe and consistent adaptation with minimal delay.}} as the safety buffer, preventing mode changes due to a \emph{"false reactive"} classification, while introducing minimal delay, which we did not observe to negatively impact the system during hardware deployment. 

\textbf{Adaptation Goals:} When manipulating an unknown payload, the robot should be in a (stiff) \texttt{reactive mode} allowing for payload estimation updates. Conversely, when a human perturbation is detected, the robot should be in a (soft) \texttt{compliant mode}, passive to the human touch.

\subsection{Variable Impedance Control}\label{sec:vic}
We achieve the desired adaptation goals by actively adjusting the damping of the controller from~\eqref{eq:passive_impedance} by formulating a variable impedance equation $\gamma_{d}(\cdot)$ based on the interaction classification output $\hat{y}_\text{int}\in[\texttt{reactive}~(0),\texttt{compliant}~(1)]$ to switch between low damping (compliant mode) and high damping (no interaction is detected) as below,
\begin{equation}
\label{eq:adaptation_law}
    \gamma_{d,k}(\hat{y}_\text{int}) = \alpha_\gamma(1+\beta_\gamma(1-\hat{y}_\text{int}))+(1-\alpha_\gamma)\gamma_{d,k-1}
\end{equation}
where $\beta_\gamma>0$ is a maximum damping magnitude increase and $\alpha_\gamma$ is the filtering parameter, ensuring no abrupt damping changes. In most experiments, we used $\beta_\gamma=1.4$, corresponding to the maximum increase in damping for the reactive mode, and $\alpha_\gamma=0.05$, though both can be tuned arbitrarily\footref{fn:alpha_beta} or set to strictly enforce passivity by regulating energy exchange rates.

\textbf{Passivity Analysis:} As studied in \cite{7560657,shao2024constraintaware}, the stability (when $F_{\rm{ext}}=0$) and passivity (when $F_{\rm{ext}}\neq0$) of a closed-loop impedance-controlled system rely on satisfying bounds on the positive rate of change (stiff $\uparrow$) of the variable impedance parameter; in our case $\dot{\gamma}_d(\cdot)>0$. By plugging Eqs.~\ref{eq:adaptation_law}, \ref{eq:passive_impedance} into Eq.~\ref{eq:dynamics} and performing the closed-loop system energy analysis, the passivity bound for Eq.~\ref{eq:adaptation_law} is, 
\begin{equation}
\label{eq:passivity_bound}
    \gamma_{d,k} \leq \gamma_{d,k-1} + \Delta t\, \frac{ \dot{x}^{\mathsf T}\gamma_{d,k-1}D(x_{k-1})\dot{x}}{ \lambda_1\mathcal{V}(x_{k-1})+\epsilon}
\end{equation}
with $\lambda_1\in\Re_+$ being the first eigenvalue of $D(\cdot)\succ  0$ and $\mathcal{V}(\cdot)$ the Lyapunov candidate function corresponding to the DS-based motion planner $f(x)$ and $\epsilon$ a small constant guaranteeing numerical stability. In brief, we can guarantee that the system remains passive when $\gamma_d(\cdot)\downarrow$. This is crucial as we do not want to break the system’s passivity when human contact is detected. To control $\gamma_d(\cdot)\uparrow$, we utilize filtered contact estimation, which applies a safety buffer, delaying the increase in impedance gain after human contact is no longer present. This preserves the bound Eq.~\ref{eq:passivity_bound} except for a $<0.03s$ instant, when a bounded energy spike is injected into the system due to the rapid increase of $\hat{y}_\text{int}\rightarrow 1$.
However, this does not happen when a human is in contact with the robot, thereby having no negative impact on the interaction. The full passivity derivation and analysis are provided on the supplementary project website\footref{website}.


\subsection{System Identification and Contact Force Estimation}\label{sec:continous_mismatch_prediction}
To identify unknown payloads, we continuously run RME~\cite{jaszczuk2025rapid} (using an adaptive $\tau_\text{ext}$ history window, $\sim250\,\text{ms}$ on average, independent of the transformer's encoders) to produce $\hat{\theta}=\{m_\text{load},\hat{r}_\text{CoM, load}\}$, representing mass and center of mass applied to the EE. Note that a different mismatch estimator can be used, but we adopt RME due to its low latency. To selectively compensate for applied payloads only, we filter $\theta$ using the interaction classification $\hat{y}_\text{int}$ as,
\begin{equation}
    \hat{\theta}_{k}= (1-\alpha_\theta) \hat{\theta}_{k-1} + \alpha_\theta \theta_\text{update}
\end{equation}
\begin{equation}
    \theta_\text{update} = \begin{cases}
        \theta_\text{RME} & \text{if}~~~\hat{y}_\text{int} = 0 \\
        \theta_{\text{RME},~k-250} & \text{otherwise\footref{fn:k250}}
    \end{cases}
\end{equation}
where, if contact is detected, we recover a past $\theta$ estimate, allowing the controller to handle interaction estimation delays, while applying a simple filter with $\alpha_\theta = 0.005$ to eliminate spikes in mass prediction at the interaction state transition\footnote{\label{fn:alpha_beta}{Filtering constants $\alpha$ can be chosen arbitrarily to regulate filtering delay and performance, based on the control loop frequency (for our experiments, control loops run at 1 kHz). $\beta_\gamma$ parameter controls the maximum increase in damping of the variable impedance controller and can be tuned arbitrarily to match the desired stiffness of the final controller.}}. Further, we compute the resulting load wrench and filtered interaction forces as
\begin{equation}
    \hat{W}_\text{load} = \begin{bmatrix}
         \hat{F}_\text{load}\\ \hat{r}_{\text{CoM, load}} \times \hat{F}_\text{load}
     \end{bmatrix},~~\hat{F}_\text{load}=m_\text{load}~g
\end{equation}
\begin{equation}
    \hat{F}_{\text{int},k}= \alpha_{F_\text{int}}\underbrace{(F_\text{ext}-\hat{F}_\text{load})}_\text{\textcolor{royalblue}{Interaction Force}} + (1-\alpha_{F_\text{int}}) \hat{F}_{\text{int},k-1}
\end{equation}
with a filter with $\alpha_{F_\text{int}} = 0.05$, which can be set arbitrarily\footref{fn:alpha_beta}.

\section{Evaluation and Experimental Validation}
\subsection{Behavior Under Contact}
The critical safety requirement is that the controller remain in \texttt{compliant mode} whenever contact actually occurs, as misclassifying an ongoing contact as \texttt{reactive mode} could cause the controller to compensate for human perturbation rather than yield to it. Table~\ref{tab:updated_results_cont} evaluates how often the controller \textbf{was correctly in compliant mode} while contacts were actually taking place. After applying the safety cap, our transformer-based method (MM-Trans.) remained compliant during contact in 95.59\% of all contact instances, compared to 73.03\% for the baseline~\cite{jaszczuk2025rapid} (also using the safety buffer). This confirms that our learned classifier significantly reduces the risk of the undesired stiff \texttt{reactive} behaviors during human touch. Comparing MM-MLP and MM-Trans. show a consistent improvement in overall contact detection from 93.01\% to 95.59\%. The lightweight transformer model's strength is in detecting load- and perturbation-induced forces that are difficult to disambiguate, thereby enhancing safety in complex interaction scenarios while maintaining real time.
\begin{table}[!t]
\renewcommand{\arraystretch}{1.25}
\centering
\caption{\% of contact instances with controller in \texttt{compliant mode}.}
\label{tab:updated_results_cont}
\begin{tabular}{ccccc}
 \toprule
\textbf{Experiment} & \textbf{Contacts} & \textbf{Baseline} & \textbf{MM-MLP} &  \textbf{MM-Trans.} \\ \hline
\rowcolor{gray!10}  Perturbation &9,461 & 91.45\%& 93.24\%&94.95\%\\ 
Perturbation (z) & 1,722& 25.09\% & 85.77\%& 93.84\% \\ 
\rowcolor{gray!10} Load + Pert. &3,381 &  81.87\% &94.20\% &95.59\% \\ 
Load + Pert. (z) &3,147  & 34.41\% & 95.07\%& 98.47\%\\ \hline
\textbf{Total}& \textbf{17,711}& \textbf{73.03\%} & \textbf{93.02\%}& \cellcolor{green!10} \textbf{95.59\%}  \\ \bottomrule
\end{tabular}
\vspace{-12pt}
\end{table}

\vspace{-5pt}
\subsection{Unknown Load Manipulation with Human Interactions}
\begin{figure}[!t]
\centering
\input{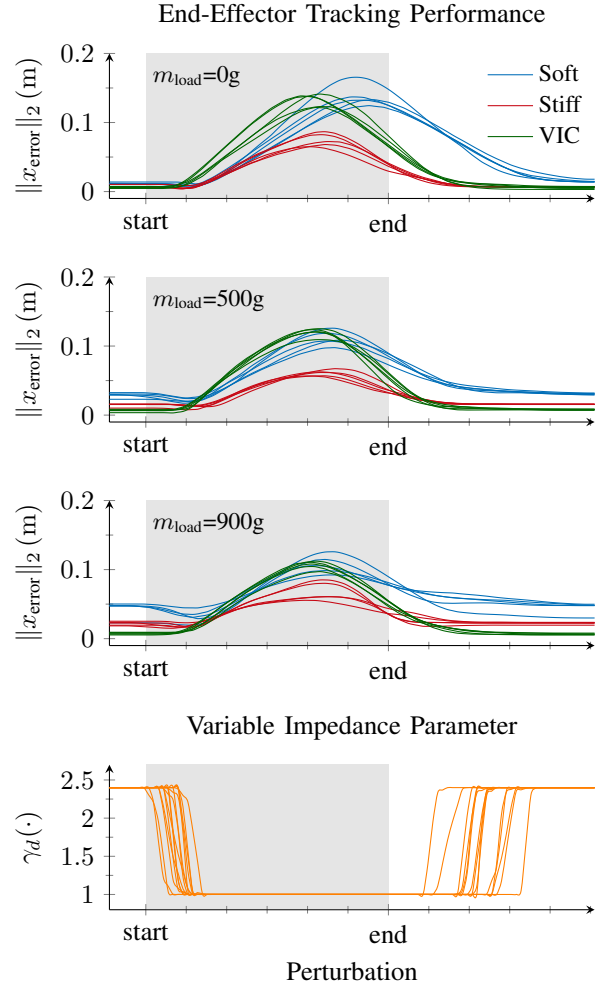}
\caption{Tracking performance of the adaptive (ours), soft and stiff controllers as the robot maintains an equilibrium point under a series of human-applied perturbations (shaded gray regions) across varying unknown payloads ($m_\text{load}$). The bottom plots illustrate the variable impedance parameter of our adaptive controller, which decreases during detected contact and increases once interaction ends. The average perturbation length was $1.245\pm0.213$ seconds over 45 trials. Ours consistently returns to the equilibrium point with the lowest residual error, while remaining compliant throughout contact.\label{fig:updated_results_error_rec}}
\vspace{-5pt}
\end{figure}

To examine the controller's accuracy during continuous adaptation, we deploy it on the Franka arm and perturb the robot as it manipulates an unknown payload. The framework distinguishes load- and interaction-induced forces online, adjusting impedance parameters and gravity compensation in real time, allowing the robot to maintain low tracking error while remaining compliant under contact. Fig.~\ref{fig:updated_results_error_rec} shows the robot tracking an equilibrium point across a series of human-applied perturbations under varying unknown loads, comparing our adaptive controller against fixed-gain stiff and soft controller baselines. After each perturbation, our controller quickly returns to the equilibrium and has lower residual error than the baselines (the average tracking errors after recovering from perturbations are shown in Table~\ref{tab:updated_results_error_rec}). The stiff controller ($\gamma_d = 2.4$) achieves relatively low tracking error, but offers no compliance during interaction, while the soft controller ($\gamma_d = 1.0$) remains compliant but recovers slowly and drifts under unknown loads. Our adaptive controller combines the benefits of both, remaining compliant during contact while precisely tracking equilibrium once the interaction ends.

\subsection{Contact Force and Intent Estimation}
\begin{figure}[!t]
\centering
\includegraphics[width=1\linewidth]{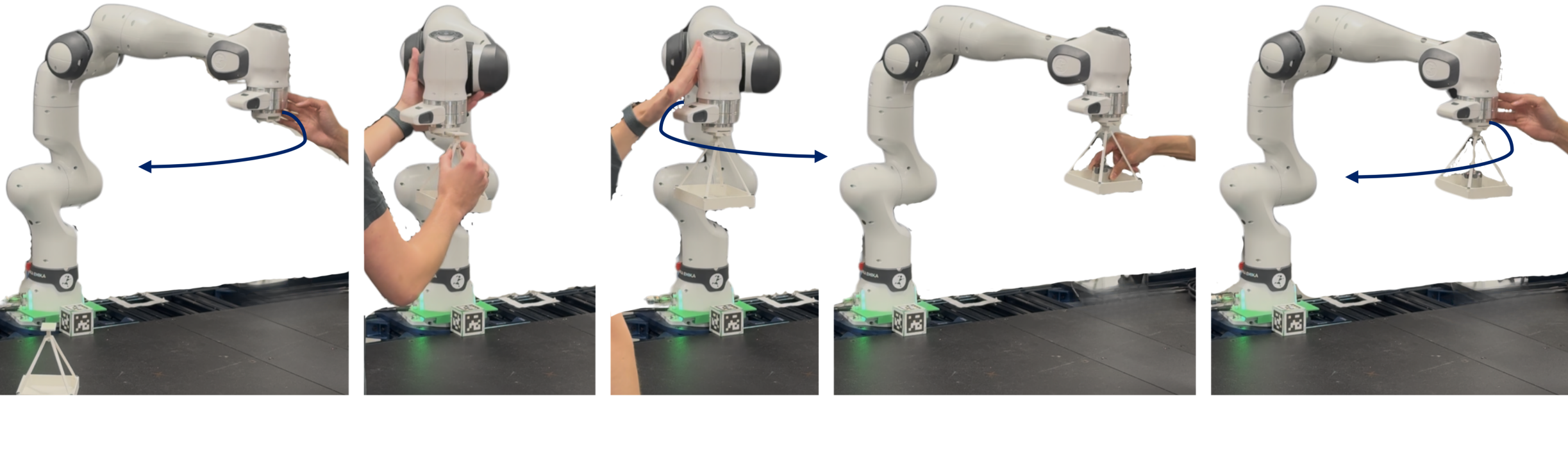}
\vspace{-20pt}
\caption{\label{fig:intent_exp}Intent estimation using \textcolor{royalblue}{interaction force} estimation. As shown, the framework provides informative estimates of interactions, enabling natural target switching upon detecting human intent.}
\vspace{-15pt}
\end{figure}

To evaluate interaction force estimation, we use $\hat{F}_\text{int}$ as the input to a Hidden Markov Model-based target belief filter inspired by ~\cite{Khoramshahi2019}, which estimates the robot's next target by evaluating the alignment of $\hat{F}_\text{int}$ with the end-effector-to-target distance vector. Further implementation details of this HMM target belief estimator are provided in the supplementary project website\footref{website}. The adaptive controller provided accurate interaction force estimation, enabling target inference and a collaborative human-robot-human transportation task, as shown in Fig.~\ref{fig:intent_exp} and in the video.

\begin{table}[!tbp]
\renewcommand{\arraystretch}{1.25}
\centering
\caption{Tracking error after recovery from perturbation/payload.}
\label{tab:updated_results_error_rec}
\begin{tabular}{c|c|c|c}
 \toprule
 \multirow{2}{*}{\makecell{Applied\\Load (g)}} & \multicolumn{3}{c}{Avg. Tracking Error $(\times10^2$ m)} \\ \cline{2-4}
  & Soft & Stiff & Adaptive (Ours) \\ \hline
\rowcolor{gray!10} 0 & 1.5$\pm$0.15 & 0.7$\pm$0.02 & \textbf{0.6$\pm$0.16} \\
500 & 3.1$\pm$0.10 &1.6$\pm$0.02&\textbf{0.8$\pm$0.07} \\ 
\rowcolor{gray!10} 900 & 4.8$\pm$0.07& 2.2$\pm$0.15& \textbf{0.6$\pm$0.09}\\ \hline
\textbf{Average}& 3.13 & 1.50 & \cellcolor{green!10} \textbf{0.67}  \\ \bottomrule
\end{tabular}
\vspace{-10pt}
\end{table}

\subsection{Bimanual Manipulation}
\begin{figure}[!t]
\centering
    \includegraphics[width=1\linewidth]{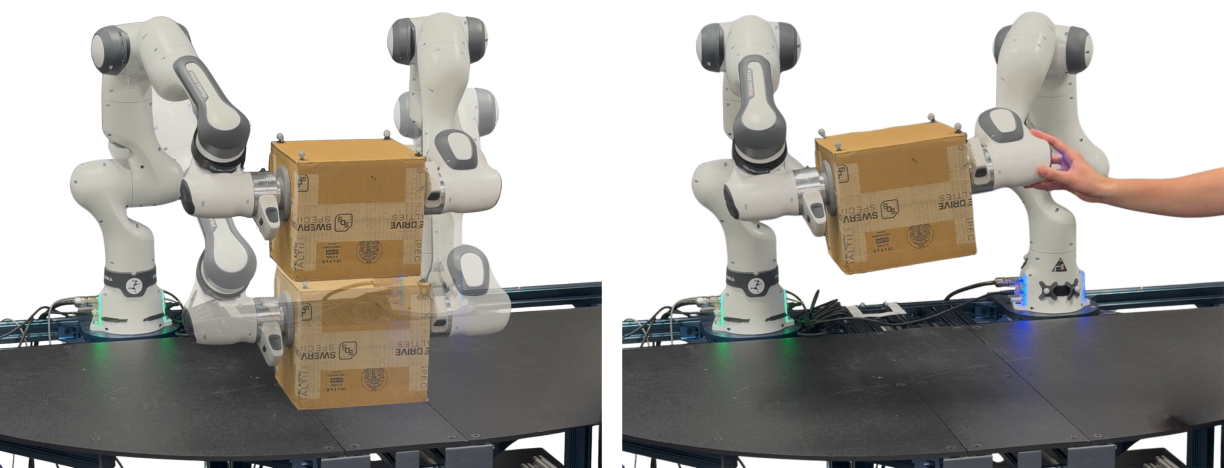}
\caption{\label{fig:bimanual}Zero-shot framework deployment on bimanual setup.}
\vspace{-5pt}
\end{figure}

To verify the framework on a more complex and untrained robotic setup, we deploy it on two Franka arms that simultaneously lift a box of \textbf{unknown mass (0.32-1\,kg)}, thereby testing adaptation within a closed kinematic chain under external perturbation. The same RME framework and interaction classifier as in the single-arm experiment are used (both robots use the more conservative of the two individual estimates); the classifier is fine-tuned only on real-robot data to improve sim-to-real transfer in an unseen setup\footnote{\label{fn:model}The model uses sim-based pretraining, but only real data for fine-tuning.}. As shown in Fig.~\ref{fig:bimanual} (and the attached video), the framework transfers zero-shot to simple bimanual tasks, with the adaptive controller reducing the steady-state tracking error along the $z$-axis from \textbf{3.34\,cm} and \textbf{1.56\,cm} for soft and stiff controllers, respectively, to \textbf{0.71 cm}, seamlessly blending adaptive and reactive behaviors.

\subsection{Dynamics Mismatch Imitation Learning Policy}
To evaluate our framework with an imitation learning policy, we deploy the pretrained EMP policy~\cite{11246142} for the pouring task with an \textbf{unknown 0.85\,kg payload}. We compare the performance of stiff ($\gamma_d = 2.4$) and soft ($\gamma_d = 1.0$) controllers against our adaptive impedance controller (with the NN classifier fine-tuned only on real-robot data to improve sim-to-real transfer to an unseen setup\footref{fn:model}). As shown in Fig.~\ref{fig:emp}, the soft controller yields to the applied payload and fails to complete the task in all three trials, diverging from the trajectory and colliding with the table. The stiff controller improves tracking; however, it fails to pour the payload in 2 trials due to collision with the pot and inability to complete the task (pouring failure), as it cannot compensate for unknown dynamics. The adaptive controller, which adjusts impedance and gravity compensation online, most closely follows the EMP trajectory (Table~\ref{tab:emp_tab}) and succeeds in all three trials, allowing the controller to safely and precisely execute the learned policy with unknown loads, while maintaining compliance to human perturbations. Note that the EMP policy, as with other pure-motion-based approaches, was not trained for unknown, heavy payload manipulation. It was trained to perform a pouring motion with the cup holding toy blocks of less than 120g. Thus, we are showing that our framework can improve the performance of motion-based imitation learning policies on OOD (out-of-distribution) dynamics.

\section{Conclusion and Future Work}
\label{sec:conclusion}
We propose a novel interaction estimation and adaptive control framework that enables collaborative robots to compensate for unknown heavy objects in real time while estimating human interaction forces, thereby enabling precise and adaptive physical interactions and outperforming single-modality baselines. We evaluated our method in a series of experiments across single and bimanual tasks, demonstrating fast and safe adaptation to constantly changing dynamics and interaction conditions, while tracking accurately. 

In our future work, we aim to improve the controller's performance by fine-tuning on more diverse datasets (across multiple locations and users) to enhance generalization and develop morphology-specific model checkpoints (single-arm, bimanual, mobile manipulation). A current limitation is that the interaction classifier assumes that load and perturbations will not occur simultaneously; when they do, the robot enters a compliant mode and delays mismatch compensation until the perturbation ends. While safe, this might compromise system performance in complex interaction scenarios, which we plan to address in our future work. We also plan to deploy this framework with additional learning-based policies, enabling it to be interactive and contact-aware around humans.

\begin{figure}[!t]
\centering
\input{TikZ_plots/emp}
\caption{EMP deployment on the \textcolor{venetianred}{soft}, \textcolor{DarkGreen}{stiff}, and \textcolor{NavyBlue}{adaptive} controllers, compared against \textcolor{orange}{nominal desired} EMP's trajectory; \textbf{\textcolor{red}{$\times$}} indicates task failure, and \protect\tikz[baseline=-0.5ex]{\protect\node[star, star points=5, star point ratio=2, fill=gray, draw=none, inner sep=0pt, minimum size=8pt]{};} indicates target/pot position.}
\label{fig:emp}
\vspace{-5pt}
\end{figure}
\begin{table}[!t]
\renewcommand{\arraystretch}{1.25}
\centering
\caption{Tracking errors evaluated with Mean Tracking Error (the mean pointwise Euclidean distance between the current and the nominal trajectory), and the Mean Dynamic Time Warping Distance~\cite{pmlr-v78-ravichandar17a} (3 trials each).}
\label{tab:emp_tab}
\begin{tabular}{cccc}
 \toprule
\textbf{Controller}  &\textbf{Soft} & \textbf{Stiff} & \textbf{Adaptive (ours)} \\ \hline
Mean TE (m) & 0.088$\pm$0.015&0.028$\pm$0.002&  \cellcolor{green!10} \textbf{0.011$\pm$0.002}\\ 
Mean DTWD (m) & 5.990$\pm$0.563&0.585$\pm$0.196 & \cellcolor{green!10} \textbf{0.194$\pm$0.030} \\ \bottomrule
\end{tabular}
\vspace{-15pt}
\end{table}

\textbf{Acknowledgment:} ChatGPT 5.1-5.2 and Claude Sonnet 5 were used to assist with code development (framework and visualization).

\vspace{-10pt}
\bibliographystyle{IEEEtran}
\bibliography{example}

\end{document}